\documentclass{elsarticle}

\usepackage{adjustbox}
\usepackage{array}
\usepackage{booktabs}
\usepackage{amsmath,amsfonts,amsthm,amssymb}
\usepackage{bm}
\usepackage{nicefrac}
\usepackage{microtype}
\usepackage{multirow}

\usepackage{enumerate}
\usepackage{titlesec}
\usepackage{listings}
\usepackage{enumitem}

\newcolumntype{L}[1]{>{\raggedright\let\newline\\\arraybackslash\hspace{0pt}}m{#1}}
\newcolumntype{C}[1]{>{\centering\let\newline\\\arraybackslash\hspace{0pt}}m{#1}}
\newcolumntype{R}[1]{>{\raggedleft\let\newline\\\arraybackslash\hspace{0pt}}m{#1}}

\newcommand{\ignorethis}[1]{}

\makeatletter
\DeclareRobustCommand\onedot{\futurelet\@let@token\@onedot}
\def\@onedot{\ifx\@let@token.\else.\null\fi\xspace}

\def\eg{\emph{e.g}\onedot} 
\def\ie{\emph{i.e}\onedot}

\def\wrt{w.r.t\onedot} 

\makeatother

\usepackage{hyperref}
\hypersetup{
    urlcolor=mydarkblue,
}
\usepackage{hyperref}
\usepackage{color,xcolor}

\makeatletter
\newif\if@restonecol
\makeatother

\usepackage[ruled,vlined]{algorithm2e}%[ruled,vlined]{
\usepackage{algpseudocode}

\def\equationautorefname~#1\null{Eq.~(#1)\null}
\newcommand{\aref}[1]{\hyperref[#1]{Appendix~\ref{#1}}} 
\newcommand{\yh}[1]{#1}
\newcommand{\yk}[1]{#1}

\journal{Pattern Recognition}

\begin{document}

\begin{frontmatter}

%% Title, authors and addresses

%% use the tnoteref command within \title for footnotes;
%% use the tnotetext command for theassociated footnote;
%% use the fnref command within \author or \address for footnotes;
%% use the fntext command for theassociated footnote;
%% use the corref command within \author for corresponding author footnotes;
%% use the cortext command for theassociated footnote;
%% use the ead command for the email address,
%% and the form \ead[url] for the home page:
%% \title{Title\tnoteref{label1}}
%% \tnotetext[label1]{}
%% \author{Name\corref{cor1}\fnref{label2}}
%% \ead{email address}
%% \ead[url]{home page}
%% \fntext[label2]{}
%% \cortext[cor1]{}
%% \affiliation{organization={},
%%             addressline={},
%%             city={},
%%             postcode={},
%%             state={},
%%             country={}}
%% \fntext[label3]{}

\title{Joint A-SNN: Joint Training of Artificial and Spiking Neural Networks via Self-Distillation and Weight Factorization}

%% use optional labels to link authors explicitly to addresses:
%% \author[label1,label2]{}
%% \affiliation[label1]{organization={},
%%             addressline={},
%%             city={},
%%             postcode={},
%%             state={},
%%             country={}}
%%
%% \affiliation[label2]{organization={},
%%             addressline={},
%%             city={},
%%             postcode={},
%%             state={},
%%             country={}}

\author{Yufei Guo\fnref{eqa}}
\author{Weihang Peng\fnref{eqa}}
\author{Yuanpei Chen}
\author{Liwen Zhang}
\author{Xiaode Liu}
\author{Xuhui Huang}
\author{Zhe Ma\corref{cor1}}
\fntext[eqa]{Equal Contributions.}
\cortext[cor1]{Corresponding author.}

\begin{abstract}
%% Text of abstract

Emerged as a biology-inspired method, Spiking Neural Networks (SNNs) mimic the spiking nature of brain neurons and have received lots of research attention. SNNs deal with binary spikes as their activation and therefore \yk{derive} extreme energy efficiency on hardware. 
However, it also leads to an intrinsic obstacle that training SNNs from scratch requires a re-definition of the firing function for computing gradient. 
Artificial Neural Networks (ANNs), however, are fully differentiable to be trained with gradient descent. 
In this paper, we propose a joint training framework of ANN and SNN, in which the ANN can guide the SNN's optimization. 
This joint framework contains two parts: First, the knowledge inside ANN is distilled to SNN by using multiple branches from the networks. 
Second, we restrict the parameters of ANN and SNN, where they share partial parameters and learn different singular weights. 
Extensive experiments over several widely used network structures show that our method consistently outperforms many other state-of-the-art training methods. For example, on the CIFAR100 classification task, the spiking ResNet-18 model trained by our method can reach to \textbf{77.39}\% top-1 accuracy with only 4 time steps.

\end{abstract}

%%Graphical abstract
%\begin{graphicalabstract}
%\includegraphics{grabs}
%\end{graphicalabstract}

%%Research highlights
%\begin{highlights}
%\item Research highlight 1
%\item Research highlight 2
%\end{highlights}

\begin{keyword}
%% keywords here, in the form: keyword \sep keyword
 Spiking Neural Networks \sep Artificial Neural Networks \sep Knowledge Distillation \sep Weight Factorization
%% PACS codes here, in the form: \PACS code \sep code

%% MSC codes here, in the form: \MSC code \sep code
%% or \MSC[2008] code \sep code (2000 is the default)

\end{keyword}

\end{frontmatter}

%% \linenumbers

%% main text
\section{Introduction}
\label{}

% The idea of Artificial Neural Networks (ANN) is highly inspired by neuroscience. 
The Spiking Neural Networks (SNNs) have been recognized as one of the next-generation neural networks \cite{maass1997networks}. 
This type of neural network is known for its bio-mimicry of the brain neurons, which utilize ``spikes" for information communication and can process data in a spatial-temporal manner \cite{roy2019nature, panda2020toward}.
SNNs imitate such spike mechanisms and thus received a lot of research attention.

Through the time dimension of spiking neurons, each of them fires a spike only if a variable called membrane potential transcends the threshold, if not, the spiking neuron would remain still. 
As a result, the activation would be either 1 (fire a spike) or 0 (remain silent), which eliminates multiplication in the neural networks and brings energy efficiency.
For example, some work shows that SNNs can save orders of magnitude energy over Artificial Neural Networks (ANNs) \cite{akopyan2015truenorth,davies2018loihi}. 
In addition, the temporal processing ability of SNNs makes them unique to ANNs in a way that learns spatial-temporal information.

Despite its energy efficiency and special spatial-temporal processing ability, training SNNs is challenging because spiking neurons fire discrete spikes whose gradients are not well-defined. 
Consequentially, the zero-but-all gradient makes it impossible to train SNNs via gradient-based optimization methods. 
To address this, various training methods have been proposed.
First, spike-timing-dependent plasticity (STDP)  \cite{bi1998synaptic} approaches \cite{mozafari2019bio,2019UnsupervisedFalez} leverage \yk{an} unsupervised learning algorithm called Hebbian learning \cite{hebb2005organization} to update the weights. STDP is inspired by biology where the spiking time difference can determine the connection of the synaptic weight. 
Yet, STDP is limited to small-scale datasets, which may be due to the lack of global information about the error. 

Second, the Surrogate Gradient (SG) tries to find an alternative function when doing back-propagation of the spiking neurons. 
Therefore, SG can be incorporated into the current gradient-based optimization framework. 
SG provides decent performance and can narrow the time steps into (\ie 5 $\sim$ 20) even on a large-scale dataset such as ImageNet~\cite{deng2009imagenet}. 
However, the SG inevitably introduces some error in calculating the gradient, though sometimes it is hard to estimate the error, which, makes the convergence unstable and slower than ANN training.

Last, the ANN-to-SNN conversion \cite{deng2021optimal,li2021free} offers another approach. It utilizes the well-trained ANN checkpoint and converts it into SNN by replacing the activation function from ReLU with spiking activation.
It provides a fast way to obtain an SNN without using gradient descent at all. 
The disadvantage of this type of method is that SNN does not have its own learned feature, rather, all SNN does is to mimic ANN. 

\begin{figure}
    \centering
    \includegraphics[width=\linewidth]{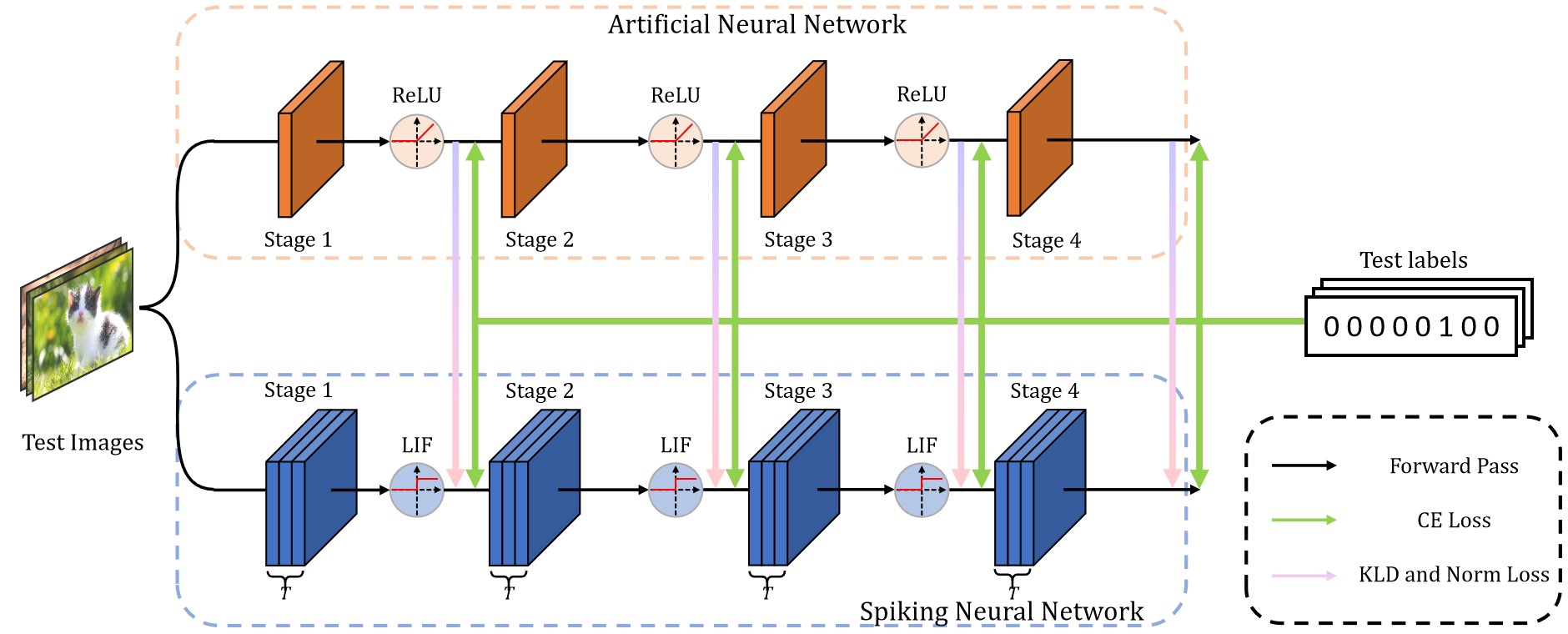}
    \caption{The overall joint training framework. \yh{To better transfer knowledge, some multiple branches with a global average pooling layer and a fully-connected layer are added inside the network. Then three kinds of distillation losses are applied to these branches: the cross-entropy (CE) loss between the output and the label, the Kullback-Leibler Divergence (KLD) loss between the ANN’s output and the SNN’s output, and the L2 loss between the output features of ANN and SNN. }}
    \label{fig_har}
\end{figure}

In this work, we propose to employ a joint training framework of both ANN and SNN. During training, the parameters of ANN and SNN are shared partially. Specifically, we apply the matrix factorization of weights via Singular Value Decomposition (SVD) and let  ANN and  SNN optimize their own singular weights but keep the singular vectors the same. 
Second, we propose adding multiple branches in the network to distill the knowledge from ANN to SNN, called self-distillation. 
Our method can be viewed as a combination of ANN-SNN conversion and SG gradient training. We do not alter the SG training algorithm in SNN, but we provide guidance during the training of SNN. 
Unlike ANN-SNN conversion, where the ANN and SNN share the same parameters, we add more degree of freedom, \ie, different singular values in parameters, but restricting them to be totally different. 
\yk{The overall workflow can be seen in \autoref{fig_har}.}
Extensive experiments over several popular network structures show that our joint training framework consistently outperforms state-of-the-art training methods. For example, on the CIFAR100 classification task, we can train a spiking ResNet-18 and achieve 77.39\% top-1 accuracy.

In summary, our contributions are three-fold:

\begin{enumerate}[leftmargin=*]
    \item We propose a joint training framework where the knowledge of ANN is consistently transferred to the SNN during training.
    \item We factorize the parameters of both ANN and SNN and restrict them to have the same singular vectors while learning different singular values. 
    \item Extensive experiments are conducted on vision benchmarks, demonstrating that our method reaches SOTA.
\end{enumerate}

The remainder of the paper is organized as follows: Section 2 provides a background for a better understanding of the proposed method. Next in Section 3, some preliminaries related to the method are introduced. In Section 4, the motivation of the paper, our ANN-SNN joint training framework, and the corresponding weight factorization method for the parameters in trained ANN and SNN are introduced in detail sequentially. Moreover, an overall algorithm pseudocode for better explaining the workflow of the proposed method is given. In Section 5, experimental settings and performance results of the proposed method and comparisons with other SoTA methods are presented. Finally, we summarize the research. 

\section{Related Work}

\subsection{Spiking Neural Networks (SNNs).} 
The \yk{SNN} as a biology-inspired method, is recognized as a potential architecture to reach high-level  artificial intelligence.  
Generally, there are two routers to obtain the high-performance SNNs: (1) converting a trained ANN to an SNN ~\citep{ rueckauer2017conversion, han2020rmp} and (2) training the SNN from scratch directly~\citep{wu2018stbp, wu2019direct}. 
The ANN-SNN conversion method trains a well-performed ANN first and then modifies the ANN ReLU activation to the SNN spike activation and reuses other parameters of the ANN to generate the same architecture-based SNN. To further improve the converted SNN accuracy, some interesting methods are proposed.
For example, in \cite{deng2021optimal} and \cite{li2021free}, the conversion error is decomposed to each layer and is reduced via calibrating the parameters. 
\yh{\cite{wang2018towards,bu2022optimal} further advance the analysis of conversion error and formulate a conversion-aware training technique for ANN.} 
Though this kind of method can obtain an SNN in a short time, getting a high-accuracy SNN requires many time steps which will increase the inference time at the same time.
Training an SNN directly from scratch can greatly reduce time steps, even less than 5~\citep{zheng2020going,guo2022reducing,guo2022imloss}, which receives more research attention recently. Recent works, such as  \citep{fang2021incorporating, kim2020revisiting}, by co-optimizing parameters, firing threshold, and leaky factor together respectively, have obtained comparable SNNs as the conversion method.
However, all these methods do not solve the non-differentiability
of the firing function and the gradient explode/vanish well, which will be introduced in detail in Section 4.1.
In this work, we focus on solving this problem.

\subsection{Knowledge Distillation.}
Knowledge distillation (KD) was originally introduced as a kind of  model compression method by \cite{hinton2015distilling}. \cite{hinton2015distilling} trains a small student model via transferring the knowledge of a rather large trained teacher model to the small one. In his work, the knowledge is defined as the teacher’s outputs after the final softmax layer. It carries richer information than one-hot labels since it can provide the inter-class similarities information learned by the teacher, which the labels do not enjoy. Following this, many works extract the intermediate representations of the teacher as the knowledge to facilitate the optimization of the student, such as intermediate feature tensors \cite{Adriana2015} or attention maps \cite{2022Paying}. Another line of KD methods
transfers the feature relationships rather than the actual features themselves~\cite{Yim_2017_CVPR}. Along with classification tasks, Many works that use the KD based learning in other tasks, e.g., action detection~\cite{zhao2022progressive}, text recognition~\cite{2021Joint}, continuous emotion recognition~\cite{2022Visual}, and object detection~\cite{tang2023task}. There are also some works that introduce the KD method in the SNN domain~\cite{Kushawaha2020}. However, they all adopt a big teacher SNN model to guide the small SNN counterpart learning. In this paper, rather than compress the SNN model, we use the KD method to solve the obstacle of the gradient exploding/varnish which avoids the good training of SNNs.

\section{Preliminary}

\subsection{Leaky Integrate-and-Fire Model}

We first introduce the notation following  \cite{li2021differentiable}. 
Throughout the paper, vectors are denoted by bold italic letters. 
For instance, $\bm{x}$ and $\bm{y}$ denote the input and target output variables. 
Matrices or tensors as clear from the text are denoted by bold capital letters (\eg $\mathbf{W}$). 
Constants are represented by small upright letters, e.g., $t$.
We can denote the time dimension and the element indices with bracketed superscripts and subscripts, respectively. For example, $\bm{u}_i^{(t)}$ means the $i$-th membrane potential at time step $t$.

In this paper, the spiking neurons we use for SNNs are the well-known Leaky Integrate-and-Fire (LIF) neuron model \cite{nahmias2013leaky}. 
The variable controlling the dynamics of LIF is called membrane potential, denoted by $\bm{u}$.
At each time step, the membrane potential is updated by
\begin{equation}
    \bm{u}^{(t+1), \text{pre}} = \tau\bm{u}^{(t)} + \bm{c}^{(t+1)}, \text{where } \bm{c}^{(t+1)} = \mathbf{W} \bm{x}^{(t+1)}, \label{eq_lif}
\end{equation}
where $\tau$ is a constant within $(0, 1)$ and controls the leakage of membrane potential. 
$\bm{c}^{(t+1)}$ is the pre-synaptic input at time step $t+1$, which is charged by input current $\bm{c}^{(t+1)}$. 
The input current can be calculated from the dot-product between the weights and the spike from the previous layer.
Once the membrane potential exceeds the firing threshold $V_{th}$, a spike will be fired from the LIF neuron, given by
\begin{equation}
    \bm{y}^{(t+1)} = 
    \begin{cases}
        1 & \text{if } \bm{u}^{(t+1), \text{pre}} > V_{th} \\
        0 & \text{otherwise}
    \end{cases}, \ \ \ 
    \bm{u}^{(t+1)} = \bm{u}^{(t+1), \text{pre}}\cdot(1 - \bm{y}^{(t+1)}) \label{eq_fire}
\end{equation}
After firing, the spike output $\bm{y}^{(t+1)}$ will propagate to the next layer and become the input $\bm{x}^{(t+1)}$ of the next layer, (note that we omit the layer index for simplicity). 

\textbf{The Classifier in SNN. }
Generally, in a neural network-based classification model, the final network output will be used to compute the softmax and predict the desired class object.
If we directly output the number of spikes to compute the probability, it will lose too much information. This is because the output of the network could be positive and negative as well. 
The spikes, however, can only output positive values.
For this reason, we choose to only integrate the network output and do not fire them across time, as did in recent practice~\cite{rathi2020dietsnn}. 
\begin{equation}
    \bm{y}_{\text{net}} = \frac{1}{T}\sum_{i=1}^{T} \bm{c}_{\text{net}}^{(t)} =  \frac{1}{T}\sum_{i=1}^{T}\mathbf{W} \bm{x}^{(t)},
\end{equation}
Then, we can compute the cross-entropy loss based on the true label and $\mathrm{Softmax}(\bm{y}_{\text{net}})$.

\section{Method}
\label{sec_method}

In this section, we introduce our ANN-SNN joint training framework. First, we discuss the motivation of this work, \ie the gradient computation problem in spiking neurons.
Then, we  present self-distilled training with multiple branches.
In the end, we present the weight factorization for the parameters in ANN and SNN.

\newcommand{\FracPartial}[2]{\frac{\partial #1}{\partial #2}}
\subsection{Motivation}
The most notorious problem in SNN training may be the non-differentiability of the firing function \autoref{eq_fire}.
To concretely discuss this problem, we denote the loss function as $L$ and calculate the gradients \wrt weights using the chain rule:
\begin{equation}
    \FracPartial{L}{\mathbf{W}} = \sum_t\FracPartial{L}{\bm{y}^{(t)}} \FracPartial{\bm{y}^{(t)}}{\bm{u}^{(t), \text{pre}}} \FracPartial{\bm{u}^{(t), \text{pre}}}{\bm{c}^{(t)}} \FracPartial{\bm{c}^{(t)}}{\mathbf{W}}.
\end{equation}
Here, the firing function (\autoref{eq_fire}) is similar to the $\mathrm{sign}$ function. The gradient of $\mathrm{sign}$ is 0 almost everywhere except for the threshold. 
Recall that the gradient descent updates the parameters by subtracting the learning rate multiplied by the gradient, therefore, the actual updates for weights would either be 0 or infinity.

As mitigation of this problem, the surrogate gradient was proposed.
When performing the forward pass, the firing function remains exactly the same, however, when performing the backward pass, the firing function becomes a surrogate function. And we can compute the surrogate gradient function.

A popular surrogate gradient may refer to the rectangular function proposed in~\cite{wu2019direct}.
Another type is called rectangular function, given by
\begin{equation}
\begin{cases}
    \FracPartial{\bm{y}^{(t)}}{\bm{u}^{(t), \text{pre}}} & =
    \gamma \max\left(0, 1 - \left|\frac{\bm{u}^{(t), \text{pre}}}{V_{th}} - 1\right|\right), \\
    \FracPartial{\bm{y}^{(t)}}{\bm{u}^{(t), \text{pre}}} & = 
    \frac{1}{a} \mathrm{sign}\left(\left|\bm{u}^{(t), \text{pre}} - V_{th}\right| < \frac{a}{2}\right).
\end{cases}
\end{equation}
Both of them have a hyper-parameter to control the width and the sharpness of the surrogate gradient.
The rectangular function will become the Straight-Through Estimator \cite{bengio2013ste} if $V_{th}=0.5, a=1$.
The triangular is computed based on the distance between the potential and threshold.

\noindent\textbf{Gradient Explode/Vanish. }
\yh{The gradient explode or vanish problem in ANN can be effectively mitigated by residual block, which contains the skip connection \cite{he2016deep}.
The skip connection can be formulated by
\begin{equation}
    \bm{y} = g(f(\bm{x}) + \bm{x}),
\end{equation}
where $g(\cdot)$ is the activation function, \eg ReLU in ANN, and $f(\cdot)$ is the convolutional layers and activation layers in the main path. Since ReLU is unbounded for the positive part, the gradient can be passed to the input of the block.
However, in the case of LIF neurons, the gradient will be reduced through SG. 
Either $\bm{u}^{(t), \text{pre}} > 2V_{th}$ or $\bm{u}^{(t), \text{pre}} < 0$
the gradient of LIF will be 0. 
Moreover, the output of $g(\cdot)$ is 0/1, which cannot effectively carry the information from the input tensor to the network output.
Thus, even using ResNet and SG the SNN still cannot be well-trained due to the potential gradient exploding/varnish. }

\subsection{Self Distillation}

Here, we propose our joint-training framework, which composes of distillation of partial weight sharing.
Let us divide the ResNet into several parts. The ResNet has a stem (the first convolution), a body, and a head (the global average pooling layers and the classifier). The body is made of 4 stages, where each stage composes several residual blocks. 
The residual blocks have two or three convolution layers and the skip connection. 
\yh{As we discussed before, even equipped with residual blocks, the SNN may still suffer from the gradient explode/vanish problem. 
Thus, we add knowledge distillation from ANN to SNN.} Particularly, the source of knowledge comes from not only the network output but also the intermediate output. 
Similar to Branchynet \cite{teerapittayanon2016branchynet} we add multiple branches inside the network. 
In total, there are four intermediate branches, which are followed right after the end of each stage of the ResNet with a global average pooling layer and a fully-connected layer.

The overall loss function for this distillation framework is composed of 4 parts:

\begin{enumerate}[leftmargin=*]
    \item Loss 1: The cross-entropy (CE) loss between the output and the label. Note that, the CE loss is introduced for both ANN and SNN, and also both the intermediate output from ANN and SNN. Thus, all branches in the framework receive direct knowledge from the labels. It can be formulated by:
    \begin{equation}
        L_{\text{CE}} = \sum_{i=1}^4CrossEntropy(\bm{y}_{\text{ANN}, i, \bm{z}}) + \sum_{i=1}^4CrossEntropy(\bm{y}_{\text{SNN}, i, \bm{z}}),
        \label{eq_l_ce}
    \end{equation}
    where both ANN and SNN have 4 stages and thus there are 8 CE losses. 
    \item Loss 2: The Kullback-Leibler Divergence (KLD) loss between the ANN's output and the SNN's output. This loss function is computed based on the softmax (\ie class probabilities), and
    ensures that the output from SNN should mimic that from ANN. 
    It can be formulated by:
    \begin{equation}
        L_{\text{KLD}} = \sum_{i}^4KLD(\mathtt{detach}(\bm{y}_{\text{ANN}, i}), \ \bm{y}_{\text{SNN}, i}).
    \label{eq_l_kld}
    \end{equation}
    Again, this loss function is applied to all branches of the network. Note that the output from ANN is detached from calculating the gradients. Otherwise, the ANN's output will be pulled over to the SNN's output, leading to an accuracy decrease. 
    \item Loss 3: The L2 loss from hints. Due to the temporal dimension and binary activation in SNN architecture, the features in ANN and the features in SNN may have different norms of magnitude. Therefore, we impose L2-norm loss between the output features of ANN and SNN, given by
    \begin{equation}
        L_{\text{Norm}} = \sum_{i}^4|| \mathtt{detach}(\bm{y}_{\text{ANN}, i})- \ \bm{y}_{\text{SNN}, i} ||_F^2,
    \label{eq_l_norm}
    \end{equation}
    Similar to KLD loss, the L2 norm loss also detaches the output from ANN from computing the gradients. 
\end{enumerate}

Together, we optimize these three loss functions simultaneously to transfer the knowledge from ANN to SNN. Remarkably, the loss function from the first and second stage's outputs provides the guiding information for shallow layers, which alleviates the gradient explode/vanish problems. The overall loss function can be written by 
\begin{equation}
    L = L_\text{CE} + \lambda_1 L_\text{KLD} + \lambda_2 L_\text{Norm}, 
\end{equation}
where $\lambda_1$ and $\lambda_2$ are the hyperparameters for controlling the distillation loss.

\subsection{Weight-Factorization Training}

The second ingredient of our joint-training framework is weight-factorization training (WFT). This method is inspired by ANN-SNN conversion. 
Conventionally, the ANN-SNN conversion directly copy-pastes the weight and bias parameters from ANN to SNN \cite{diehl2016conversion}. 
This method is only effective when the time steps in SNN are large enough. \yh{However, in our training framework the SNN is trained with less than 10 time steps. Thus if we keep the parameters in the ANN and the SNN identical, the performance of the SNN is limited. }
Recently, \cite{li2021free} suggests that a proper \emph{calibration} of the weights in ANN is needed for conversion to SNN \yh{in extremely low time steps}.  
This finding inspires us to develop a \textbf{partial weight-sharing} regime for the joint training of ANN and SNN. 
% Specifically, we apply the Singular Value Decomposition (SVD) to the weights parameters and share the eigenvectors for ANN and SNN. 
\yh{In order to do so, we apply the Singular Value Decomposition (SVD) to the weights parameters and share the eigenvectors for ANN and SNN while keeping the eigenvalues optimized at the discretion of the ANN and SNN separately. 
Intuitively,} let us first consider a simple case: a fully-connected layer $\mathbf{W}\in\mathbb{R}^{c_{in}\times c_{out}}$. We can apply the SVD to the weight matrix, given by
\begin{equation}
    \mathbf{W} = \mathbf{U\Sigma V}^\top = 
    \begin{bmatrix}
    \bm{u}_1 & \bm{u}_2 & \cdots & \bm{u}_{c_{in}}
    \end{bmatrix}
    \begin{bmatrix}
    \sigma_1 & 0 & \cdots &  0 \\
    0 & \sigma_2 & \cdots &  0 \\
    \vdots & 0 & \ddots &  0 \\
    0 & 0 & \cdots &  \sigma_r \\
    \end{bmatrix}
    \begin{bmatrix} 
    \bm{v}_1 \\ \bm{v}_2 \\ \vdots \\ \bm{v}_{c_{out}}
    \end{bmatrix},
\end{equation}
where $\bm{u}_i\in\mathbb{R}^{c_{in}}$ and $\bm{v}_i\in\mathbb{R}^{c_{out}}$ are the $i$-th eigenvectors of $\mathbf{WW}^\top$ and $\mathbf{W}^\top\mathbf{W}$, respectively. $\sigma$ is the singular value and $r$ is the number of singular value. 
SVD can decompose weights into:
\begin{equation}
    \mathbf{W} = \sum_{i=1}^r \sigma_i \bm{u}_i\bm{v}_i^\top.
\end{equation}
In other words, SVD can decompose the weight matrix into a weighted sum of several rank-1 matrices. In our WFT, we let SNN and ANN have the same eigenvectors\yh{ but different eigenvalues, given by
\begin{equation}
    \mathbf{W}_{\text{ANN}} = \mathbf{U\Sigma}_{\text{ANN}}\mathbf{V}, \ \ \ \mathbf{W}_{\text{SNN}} = \mathbf{U\Sigma}_{\text{SNN}}\mathbf{V}.
    \label{eq_wft}
\end{equation}
%As a result, the input activation is projected into the same plain with different factors. 
% However, the singular values for ANN and SNN are not shared. They can learn it separately. 
In this case, the SNN and ANN do not share parameters completely, providing more degree of freedom compared to the ANN-SNN conversion works. 
In our implementation, we re-parameterize the weights using Eq.~(\ref{eq_wft}) and directly update $\mathbf{U,V,\Sigma}_{\text{ANN}},$ and $\mathbf{\Sigma}_{\text{SNN}}$ with gradient descent}. 
The initial values are determined by applying the SVD for the initialized weights. The singular values in the ANN and SNN are also initialized to the same. 
We put the overall algorithm pseudocode in \autoref{alg:1}.

\begin{algorithm}[t]
    \caption{A joint training framework for ANN and SNN}
    \label{alg:1}
    \KwIn{ANN/SNN to be trained; Training dataset, total training epoch $E$, the distillation hyper-parameter $\lambda_1$ and $\lambda_2$\yk{;}}
    Initialize the weight parameters $\mathbf{W}$\;
    Apply the SVD to $\mathbf{W}$ and obtain $\mathbf{U,V,\Sigma}_{\text{ANN}},$ $\mathbf{\Sigma}_{\text{SNN}}$\;
    \For{all $e=1,2,\dots, E$-th epoch}
            {
            Compose the weights: $\mathbf{W}_{\text{ANN}} = \mathbf{U\Sigma}_{\text{ANN}}\mathbf{V}$, $\mathbf{W}_{\text{SNN}} = \mathbf{U\Sigma}_{\text{SNN}}\mathbf{V}$ \;
            Get all outputs $\bm{y}_{\text{ANN}, i}$, $\bm{y}_{\text{SNN}, i}$\;
            Compute CE loss $L_{\text{CE}}$ (cf. \autoref{eq_l_ce}) \;
            Compute KLD loss $L_{\text{CE}}$ (cf. \autoref{eq_l_kld}) \;
            Compute Norm loss $L_{\text{CE}}$ (cf. \autoref{eq_l_norm}) \;
            Back-propagation and compute the gradients: $\nabla_{\mathbf{U}}L, \nabla_{\mathbf{V}}L, \nabla_{\mathbf{\Sigma}_{\text{ANN}}}L, \nabla_{\mathbf{\Sigma}_{\text{SNN}}}L$\;
            Descend loss function and update weights\;
            }
            \textbf{return} trained SNN and trained ANN.
\end{algorithm}

\subsubsection{Memory Complexity}
Our method does not largely increase the memory of weights training when compared to the separate training baseline, \ie optimize $\mathbf{W}_{\text{ANN}}$ and $\mathbf{W}_{\text{SNN}}$ separately. 
The number of elements for our method is $c_{in}^2 + c_{out}^2+2r$ where $r = \min(c_{in}, c_{out})$, while the number of elements for the baseline is $2c_{in}c_{out}$.
For most convolutional layers in ResNets, the $c_{in}$ is the same with $c_{out}$, therefore, our method only has $2c_{in}$ additional parameters, with the same order of magnitude with bias parameters. 
For the downsampling layers, we have $c_{out}=2c_{in}$ and thus our method has 20\% more parameters, which is affordable considering that there \yk{are} only 3 downsampling layers in ResNets.

\subsubsection{Generalization to Convolutional Layers}

For Conv layers, the weights are 4D tensors, where kernel sizes add up to two additional dimensions. In our implementation, these two dimensions do not affect our implementations. We can apply the same SVD decomposition in every entry of the kernel. As a consequence, all layers can be implemented with WFT.

\subsection{Extending to Other Networks }
\yk{In addition to the ResNet we discussed before, our joint learning framework can be applied to other network architectures like VGG-series~\cite{simonyan2014very}. For these modern convolutional architectures, the network is usually partitioned into four different stages, each stage corresponding to a downsample of the input feature map and an increase of the channel numbers. We add the distillation loss function to the exit of these four stages. Thus, our method can seamlessly incorporate most convolutional networks. }

\section{Experiments}
\label{sec_exp}

In this section, we demonstrate the effectiveness and efficiency of our algorithms. 
We first briefly illustrate the implementation details of our experiments and then compare the results with existing state-of-the-art. 
We also provide ablation studies.

\subsection{Implementation Details}

\begin{table}[tphb]
\caption{Comparison between our algorithm and existing algorithms on CIFAR-10/100 datasets \yk{with ResNets}. Our method improves network performance across all tasks. }
\label{tab_cifar}
\begin{center}
\resizebox{\textwidth}{!}{
\begin{tabular}{llccc}
\toprule
\multicolumn{1}{c}{\bf Dataset}  &\multicolumn{1}{c}{\bf Method} &\multicolumn{1}{c}{\bf Architecture} &\multicolumn{1}{c}{\bf Time Steps} &\multicolumn{1}{c}{\bf Accuracy}\\ 
\midrule
\multirow{16}{*}{CIFAR10}   
    &Hybrid training \cite{rathi2020enabling} &ResNet-20 &250 &92.22\\
    &\yh{TTBR} \cite{meng2022training} & ResNet-18 & 64 & 95.04\\
    &TSSL-BP \cite{zhang2020temporal} &CIFARNet &5 &91.41\\
    & TET \cite{deng2022temporal} & ResNet-19 & 4 & 94.44 \\
    & RecDis-SNN \cite{GuoTCZLMH22} & ResNet-19 & 2 &  93.64 \\
    \cmidrule{2-5}
    & \multirow{2}{*}{Real Spike~\cite{guo2022real}} & \multirow{2}{*}{ResNet-20} & 4 &  92.53 \\
    & & & 2 &  90.47\\  
    \cmidrule{2-5}
    & \multirow{2}{*}{Dspike \cite{li2021differentiable}} & \multirow{2}{*}{ResNet-18} & 4 & 93.66 \\
    & & & 2 &  93.13\\    
    \cmidrule{2-5}
    & \multirow{2}{*}{STBP-tdBN \cite{zheng2020going}} & \multirow{2}{*}{ResNet-19} & 4 & 92.92 \\
    & & & 2 &  92.34\\
    \cmidrule{2-5}
    & \multirow{4}{*}{\bf Joint A-SNN (Ours)} & \multirow{2}{*}{ResNet-18} & 4 & \bf 95.45 \\
    & & & 2 & \bf 94.01\\
    \cmidrule{3-5}
    & & \multirow{2}{*}{ResNet-34} & 4 & \bf 96.07 \\
    & & & 2 & \bf 95.13\\
    
    % \cline{2-6}
    % &ANN* &ANN &ResNet-19 &1 &94.97\\
\midrule
    {\multirow{11}{*}{CIFAR100}}  
    &\yh{TTBR} \cite{meng2022training} & ResNet-18 & 64 & 78.45\\
    &Diet-SNN \cite{rathi2020dietsnn} &ResNet-20  &5  & 64.07\\
    & TET \cite{deng2022temporal} & ResNet-19 & 4 & 74.47 \\
    & RecDis-SNN \cite{GuoTCZLMH22} & ResNet-19 & 4 & 74.10 \\
    \cmidrule{2-5}
    & \multirow{2}{*}{Real Spike~\cite{guo2022real}} & \multirow{2}{*}{ResNet-20} & 4 &  64.87 \\
    & & & 2 &  63.40\\  
    \cmidrule{2-5}
    & \multirow{2}{*}{Dspike \cite{li2021differentiable}} & \multirow{2}{*}{ResNet-18} & 4 & 73.35 \\
    & & & 2 &  71.68\\     
    % \cmidrule{2-5}
    % & \multirow{2}{*}{Baseline (Ours)} & \multirow{2}{*}{ResNet-18} & 4 & 95.45 \\
    % & & & 2 & 94.01\\
    \cmidrule{2-5}
    & \multirow{4}{*}{\bf Joint A-SNN (Ours)} & \multirow{2}{*}{ResNet-18} & 4 & \bf 77.39 \\
    & & & 2 & \bf 75.79\\
    \cmidrule{3-5}
    & & \multirow{2}{*}{ResNet-34} & 4 & \bf 79.76 \\
    & & & 2 & \bf 77.11\\
\bottomrule
\end{tabular}}
\end{center}
\end{table}

\begin{table}[t]
\caption{\yk{Comparison between our algorithm and existing algorithms on CIFAR-10/100 datasets with VGGs.} }
\label{tab_cifar_vgg}
\begin{center}
\resizebox{\textwidth}{!}{
\begin{tabular}{llccc}
\toprule
\multicolumn{1}{c}{\bf Dataset}  &\multicolumn{1}{c}{\bf Method} &\multicolumn{1}{c}{\bf Architecture} &\multicolumn{1}{c}{\bf Time Steps} &\multicolumn{1}{c}{\bf Accuracy}\\ 
\midrule
\multirow{4}{*}{CIFAR10}   
    &Hybrid training \cite{rathi2020enabling} &VGG-16 &100 &91.13\\
    & PLIF \cite{fang2021incorporating} & VGG-11 & 8 & 93.50 \\ 
    & \yh{Temp Prune} \cite{chowdhury2022towards} & VGG-16 & 1  & 93.05\\
    \cmidrule{2-5}
    & \yk{\multirow{1}{*}{\bf Joint A-SNN (Ours)}} & \yk{\multirow{1}{*}{VGG-16}} & \yk{1} & \yk{\bf 93.79 }\\
\midrule
    {\multirow{4}{*}{CIFAR100}}  
    & Hybrid training \cite{rathi2020enabling} &VGG-11 & 125 &67.87\\
    & \yh{Temp Prune} \cite{chowdhury2022towards} & VGG-16 & 1  & 70.15\\
    \cmidrule{2-5}
    & \yk{\bf Joint A-SNN (Ours)} & \yk{VGG-16} & \yk{1} & \yk{\bf 74.24} \\
\bottomrule
\end{tabular}}
\end{center}
\end{table}

We use ResNet-18 and ResNet-34 as our selected architectures for experiments. We also use VGG-16 \cite{simonyan2014very} in some cases. 
The distillation hyper-parameters are set to $\lambda_1=1$ and $\lambda_2 = 0.3$.
We use Adam optimizer \cite{kingma2014adam} with a learning rate of $1e-3$. The learning rate is decayed using the cosine annealing strategy \cite{loshchilov2016sgdr}.
The weight decay is set to $5e-4$ and we train all models for 300 epochs and report the test accuracy in the last epoch.
We run the model with 4 GTX 1080Tis. 
We verify our models on CIFAR-10, CIFAR-100 \cite{cifar}, as well as Tiny-ImageNet \cite{deng2009imagenet}.  \yh{We encode the images to spike using the first layer in the SNN, as adopted in recent works\cite{guo2022real,deng2022temporal}.} We also give an introduction for each dataset:

\textbf{CIFAR10. } CIFAR-10 contains 50k $32\times 32$ training images and 10k test images. 
It is divided into 10 classes. 
The network kernel sizes and strides of neural networks are adjusted to fit the size of this dataset.

\textbf{CIFAR100. } CIFAR-100 consists of 50k $32\times 32$ training images and 10k test images. 
It is divided into 100 classes. 
Similarly, the network kernel sizes and strides of neural networks are also adjusted to fit the size of this dataset.

\textbf{Tiny-ImageNet. }
Tiny-ImageNet is the modified subset of the original ImageNet dataset \cite{deng2009imagenet}. Here, there are 200 different classes of ImageNet dataset, with 100k training and 10k validation images. The resolution of the images is 64×64 pixels.

\subsection{Comparison with State of the Art}
In this section, we compare our method with the existing state-of-the-art works.
We first focus on the evaluation of SNNs on CIFAR-10 and CIFAR-100 datasets.
We adopt Hybrid training, TTBR, Temp Prune, Diet-SNN, STBP-tdBN, TSSL-BP, PLIF, Dspike, Real Spike, RecDis, and TET as our comparison (see reference in the results table) and summarize the accuracy comparison in \autoref{tab_cifar}.
For the CIFAR-10 dataset, the highest accuracy from prior work is 94.44 with ResNet-19. While our joint-training method achieves 1\% absolute improvements with the same time steps based on the ResNet-18. And the ResNet-19 has 2x channel numbers and more than 10x computation cost than ResNet-18. It is also worth noting that with only 2 time steps, our method can also outperform the Real Spike and the Dspike with 4 time steps with 1.48\% and 0.35\% accuracy, respectively. These comparison results clearly show the efficiency and effectiveness of our method.

For the CIFAR-100 dataset, the TET \cite{deng2022temporal} and Dspike \cite{li2021differentiable} achieve the best accuracies, which are 73.35 and 74.47, respectively. 
Our joint training method greatly improves the accuracy to 77.39, nearly 3\% improvement.
Notably, the SNN trained by our method with even half-time steps (\ie, 2) can obtain 75.8\% accuracy, outperforming the state of the arts. 

Moreover, we train a 34-layer network with our joint-training framework, pushing the limit of SNN higher. 
For example, on CIFAR-10, our ResNet-34 reaches 96.07\% accuracy.
As for the CIFAR-100 dataset, the ResNet-34 scores a 79.8\% accuracy, a result close to 80\%.
\yk{In addition to the ResNet family architecture, we show that our framework can be applied to other networks like VGG-16. We run SNN with only 1 time step in order to compare the state-of-the-art Temporal Pruning~\cite{chowdhury2022towards}. We list the results in \autoref{tab_cifar_vgg}, where we can find our method outstrips the other SNN works tested with VGG-16. Our method achieves a much higher improvement on the CIFAR-100 dataset (4\% compared to \cite{chowdhury2022towards}). }

\begin{table}[t]
\caption{Comparison between our algorithm and existing algorithms on Tiny-ImageNet datasets. Our method improves network performance across all tasks. }
\label{tab_tiny}
\begin{center}
\resizebox{\textwidth}{!}{
\begin{tabular}{llccc}
\toprule
\multicolumn{1}{c}{\bf Dataset}  &\multicolumn{1}{c}{\bf Method} &\multicolumn{1}{c}{\bf Architecture} &\multicolumn{1}{c}{\bf Time Steps} &\multicolumn{1}{c}{\bf Accuracy}\\ 
\midrule
\multirow{4}{*}{Tiny-ImageNet}   
    &AGC \cite{kundu2021spike} & VGG-16 & 150 &51.92\\
    &DCT-SNN \cite{garg2021dct} &VGG-13 & 125 & 52.43\\
    \cmidrule{2-5}
    & \multirow{2}{*}{\bf Joint A-SNN (Ours)} & \multirow{2}{*}{VGG-16} & 4 & \bf 55.39 \\
    & & & 2 & \bf 53.91\\
\bottomrule
\end{tabular}}
\end{center}
\end{table}

Next, we conduct experiments on the Tiny-ImageNet dataset in \autoref{tab_tiny}, which is a more complex dataset than CIFAR, to verify our method. There aren't many baselines on this dataset, yet our method still achieves higher accuracy with 4 or even 2 time steps. This shows the ability of our method to handle the large-scale dataset.

\begin{table}[t]
\caption{Ablation results for different design choices. }
\label{tab_ablation}
\begin{center}
% \resizebox{\textwidth}{!}{
\begin{tabular}{cccccccc}
\toprule
\multicolumn{5}{c}{\bf Design Choices} & &\multirow{2}{*}{\bf Accuracy} & \multirow{2}{*}{\bf \yk{Time}}\\
\cmidrule{1-5}
{ w./ ANN}  & { CE Loss} &{ KLD Loss} &{ Norm Loss} &{ WFT} && \\ 
\midrule
 & \checkmark & & & && 92.39  & 26.7s\\
\checkmark & & \checkmark & & && 93.12 & 31.6s\\
\checkmark & \checkmark & \checkmark & & && 94.09 & 34.8s \\
\checkmark & \checkmark & \checkmark & \checkmark & && 94.52 & 34.9s\\
\checkmark &  &  & & \checkmark && 93.69 & 32.2s \\
\checkmark & \checkmark & \checkmark & \checkmark & \checkmark&& \bf 95.45 & 36.7s\\
\bottomrule
\end{tabular}
\end{center}
\end{table}

\subsection{Ablation Studies}

In this section, we conduct ablation studies on the design choice of our joint training frameworks. 
Here, we have several baselines: 
(1) Independent trained SNN without any distillation loss function; (2) trained SNN with vanilla KLD loss from trained ANN (similar to Knowledge Distillation \cite{hinton2015distilling}); (3) jointly trained ANN and SNN with different choices of loss functions, including CE Loss, KLD Loss, Norm Loss; (4) jointly trained ANN and SNN with weight-factorized training yet without distillation. \yk{Here, we summarize these cases together with final test accuracy and training time cost for 1 epoch running. 
Note that here we use ResNet-18 for the CIFAR-10 dataset. 
The results are put in \autoref{tab_ablation}, from which we can see that the standard training of SNN is 92.4\%, a similar accuracy level to existing works. Because standard training does not use the knowledge from ANN, the training is the fastest among other approaches, amounting to 26.7 seconds per training epoch.
If we choose to use Knowledge Distillation \cite{hinton2015distilling} (the second row), the performance would boost to 93.1\%, which is a moderate improvement (0.7\%). It also leads to a higher training time (increased by 5 seconds). 
Using the joint training with CE and KLD loss, we get a 1.7\% accuracy improvement.
Moreover, with the L2 norm loss, one even further gets another performance lift, amounting to 94.5\% final accuracy. These two choices also increase the a bit training time but get significantly higher accuracy.
Finally, we verify the WFT mechanism, which obtains 93.7\% accuracy.
This result proves that near-ANN parameters may help the optimization with SNN.
Together, these parts all contribute to the final accuracy, resulting in 95.45\% accuracy. }

We also compare the ANN performance here. Since WFT will also affect the training convergence of ANN, here we report how ANN's accuracy \yk{changes} after using WFT.
Our baseline result is 95.61\%. Using the CE loss, we add multiple branches to ANNs, which also helps the optimization, and get 95.98\% accuracy. 
Finally, using WFT the accuracy of ANN is slightly affected (95.74\%).

\subsection{\yk{Energy} Efficiency}

% \begin{table}[]
%     \centering
%     \caption{Training cost comparison across vanilla ANN, vanilla SNN, and our joint training framework, including training GPU memory usage, and training time per epoch. }
%     \begin{tabular}{l c c c c c }
%     \hline
%     \textbf{Method}  & \bf T & \bf Training GPU Mem. & \bf Training Time \\
%     \hline
%      % ANN & N/A & 48.78 MB\\
%      SNN \cite{li2021differentiable} & 2 & - &  \\
%      SNN \cite{li2021differentiable} & 4 & - & - \\
%      A-SNN (Ours) & 2 & - & 35s \\
%      A-SNN (Ours) & 4 & - & \\
%      \hline
%     \end{tabular}
%     \label{tab_traincost}
% \end{table}

\begin{table}[t]
    \centering
    \caption{Energy \yk{estimation} of ANN and SNNs \yk{of computation}.}
\begin{tabular}[b]{l c c c c c }
\hline
\textbf{Method} & \textbf{Model} & \textbf{Acc.} & \textbf{\#Add.} & \textbf{\#Mult.} & \textbf{Energy}\\
\hline
ANN (ours) & Res-18 & 95.74 & 187M & 187M &  1.03$J$ \\
\hline
SNN~\cite{rathi2020dietsnn}(T=5) & Res-20 & 92.70 & 142M & 8.80M & 168$mJ$\\
SNN (\cite{li2021differentiable}, T=2) & Res-18 & 93.13 & 71.2M & 3.52M & \textbf{80.3}$mJ$ \\
SNN (ours, T=2) & Res-18 & 94.01 & 73.4M & 3.52M & 82.2$mJ$ \\
\hline
\end{tabular}
    \label{tab_energy}
\end{table}

In this section, we measure the \yk{hardware efficiency of our proposed framework during inference. }
\yk{To measure the inference cost, we measure the energy cost of both computation and data movement. 
It is worthwhile to highlight that in some neuromorphic hardware, the data movement cost could be negligible due to the in-memory computing design \cite{davies2018loihi}. Yet in the digital circuit, the data accessing energy cannot be ignored. Hence, we discuss the energy cost of computation and data movement separately in this section.  
}  
%It is worthwhile to highlight that our framework can discard the ANN part after training and fuse the compose of the weight in SNN so that it equals a vanilla SNN in terms of model size and capacity. 

\noindent\yk{\textbf{Computational Energy Cost.} }The dot-product in ANN is equal to performing a multiply-accumulate (MAC) operation, composed of an addition and a multiplication. 
\yh{As for SNN, the binary nature of spike-based computation can eliminate the multiplication for all layers except for the first layer. }
Moreover, if the spike is not fired, the hardware can even avoid this addition with sparse computation. 
Based on this, we estimate the energy consumption following \cite{rathi2020dietsnn,horowitz20141}, a technique using 45nm CMOS technology. The MAC operation in ANN costs 4.6$pJ$ energy and the accumulation in SNN costs 0.9$pJ$ energy. Based on this, we compute the energy cost and put it in \autoref{tab_energy}. Our network only costs 82.2$mJ$ for a single forward and consumes 12.5$\times$ lower energy compared with ANN. 
Compared to existing \cite{li2021differentiable}, our model consumes slightly higher energy, amounting to 1.9$mJ$.
Nevertheless, our sacrifices only trivial energy for a large accuracy improvement over \cite{li2021differentiable} (0.9\% accuracy uplift).

\noindent\yk{\textbf{Data Movement Energy Cost.} We also estimate the energy cost of data movement in the case of digital circuit deployment. We follow that moving 64 bits of data from the cache costs $10-100pJ$ using 45nm CMOS technology too~\cite{horowitz20141}. For ANN, we calculate the total data moved as two times of activation sizes from all layers, based on that every feature map will be moved from the computational unit to the cache after the computation of the previous layer and moved from the cache to the computational unit again for the computation of the present layer. For SNN, we multiply the activation sizes by $T$ times due to the movement of membrane potential, as well as the output spikes number. Note that one spike is counted as 1 bit while one activation or one membrane potential is counted as 32 bits. 
We put the summary of data movement cost in \autoref{tab_energy_dm} where minimum energy is calculated by 10$pJ$ per 64-bit movement and maximum energy is calculated by 100$pJ$ per 64-bit movement. We can find that SNN indeed costs a slightly higher data movement energy than ANN. However, combing both data movement energy and computational energy still exhibits better energy efficiency in SNNs. 
}

\begin{table}[t]
    \centering
    \caption{\yk{Energy estimation of ANNs and SNNs of data movement}.}
\begin{tabular}[b]{l c c c c c }
\hline
\textbf{Method} & \textbf{Model} & \textbf{Acc.} & \bf Volume & \textbf{Min. Energy} & \bf Max. Energy\\
\hline
ANN (ours) & Res-18 & 95.74 & 6.234 MB & 1.022$mJ$ &10.22$mJ$\\
\hline
% SNN (\cite{li2021differentiable}, T=2) & Res-18 &93.13  \\
SNN (ours, T=2) & Res-18 &\bf 94.01 & 12.506 MB & 2.062$mJ$ & 20.62$mJ$\\
\hline
\end{tabular}
    \label{tab_energy_dm}
\end{table}

\section{Conclusion}
In this paper, we have introduced the joint-training framework of ANN and SNN. 
Our framework consists of two core ingredients, the first is self-distillation from multiple branches, and the second is weight-factorized training assisted by the Singular Value
Decomposition. 
Both of them aim to provide auxiliary information during the training of SNNs to circumvent the problem of gradient vanishing/exploding. 
Our method can be viewed as a special case of ``hybrid learning", \ie, using ANN for SNN training.
We have demonstrated that our method brings a large improvement over the original baseline and the existing state of the arts on several widely used network structures.
We hope this framework will enable more advanced neuromorphic intelligence.
\yh{However, since ANNs are experts in spatial feature processing but poor at temporal ones, depending on the guide of ANNs, our framework will limit the spatial-temporal processing ability of SNNs to some extent.
For future work, we will further study the combination manner of ANNs and SNNs to optimize the underlying code and retain the spatial-temporal processing ability of SNNs.}

\section*{Acknowledgment}

This work is supported by grants from the National Natural Science Foundation of China under contracts No.12202412 and No.12202413.

\bibliographystyle{elsarticle-num}
\bibliography{ref}

% \end{thebibliography}
\end{document}